\documentclass[10pt,twocolumn]{article}

\usepackage[margin=0.75in]{geometry}
\usepackage{amsmath,amssymb}
\usepackage{booktabs}
\usepackage{graphicx}
\usepackage{tikz}
\usetikzlibrary{arrows.meta,positioning,shapes.geometric,fit,backgrounds}
\usepackage[hidelinks]{hyperref}
\usepackage{caption}
\usepackage{xcolor}
\usepackage{mdframed}
\usepackage[numbers,sort&compress]{natbib}
\usepackage{eso-pic}
\usepackage{transparent}

\definecolor{cur}{HTML}{3b6ea5}
\definecolor{der}{HTML}{5a9367}
\definecolor{inf}{HTML}{c9a94e}
\definecolor{warn}{HTML}{b5451f}

\newmdenv[linecolor=cur,linewidth=0.8pt,roundcorner=3pt,
  backgroundcolor=cur!5,skipabove=6pt,skipbelow=6pt,
  innertopmargin=5pt,innerbottommargin=5pt]{keybox}

\title{\bf Comparative Approaches to Agent Retrieval over Large Skill Libraries}
\author{
  \begin{tabular}[t]{c}
    \textbf{Indivara Kolluru} \\
    Praetorian \\
    \texttt{ikolluru@andrew.cmu.edu}
  \end{tabular}
  \hspace{2em}
  \begin{tabular}[t]{c}
    \textbf{Nathan Sportsman} \\
    Praetorian \\
    \texttt{nathan.sportsman@praetorian.com}
  \end{tabular}
}
\date{\today}

\begin{document}
\twocolumn[
  \begin{@twocolumnfalse}
  \maketitle
  \begin{center}
  \begin{minipage}{0.85\textwidth}
  \begin{center}\large\bfseries Abstract\end{center}
  \normalsize
  \noindent Agents backed by large skill libraries must decide which skills to load and in what order. Loading the entire library into context is expensive and provides no structure for autonomous sequencing. We study two systems for this problem over a corpus of 690 skills: a hybrid ranker combining lexical and dense-embedding retrieval for sparse, on-demand loading, and a typed knowledge graph encoding workflow relations such as prerequisites, data flow, and ordering. On a set of 117 realistic, non-echoing queries, the hybrid ranker retrieves the correct skill within the top five in $73.5\% \pm 8.0$ of cases, leaving roughly a quarter of queries unserved. When used as the design intended (substituting graph neighbours for additional ranked results at matched token budget), the graph is significantly worse ($-11.2$ points, $p = 0.0007$). Its LLM-generated edge layer adds nothing over neighbours obtained free from a local embedding pass, and $73\%$ of the queries the ranker misses are not reachable through the graph at all. We attribute this to a pre-filter topology bound. Because the graph's candidate edges are drawn from the same embedding neighbourhood the ranker already searches, $98.6\%$ of typed edges connect skills the ranker had already surfaced together. The graph can enrich relation semantics but cannot extend retrieval reach. We further show that evaluating on author-written queries overstates hit@5 by up to 44 points, which would have hidden these results entirely. Our contribution is a mechanistic account of why added structure does not improve retrieval over a strong ranker, and identify the conditions under which adding structural interdependence into the retrieval is optimal.
  \end{minipage}
  \end{center}
  \vspace{1.5em}
  \end{@twocolumnfalse}
]
\thispagestyle{empty}
\clearpage

\section{Introduction}

Modern AI agents increasingly rely on libraries of reusable \emph{skills}: self-contained instruction files, each describing how to perform one capability, loaded into context when needed. As libraries grow, and ours holds $690$ skills across five departments, two questions arise: which skills to load, and in what order.

The naive answer to the first question is to load everything, which in our corpus costs ${\sim}46.9$k tokens per task, paid on every invocation regardless of need. Any serious system must beat that cost, and doing so motivated this work. Our effort produced two systems aimed at the same goal of getting an agent the right skills without loading the library. The first, and the one we set out to build, is a \textbf{typed knowledge graph} over the library, encoding workflow structure that a flat ranked list cannot represent: which skill is a prerequisite for which, whose output feeds whose input, what comes before what. The second is a \textbf{hybrid ranker}: BM25 lexical scoring fused with dense embedding similarity, serving sparse on-demand retrieval. Both are outputs of one investigation, and this paper reports what we learned by building each and then comparing them.
\label{sec:background}

We situate both systems in the loading problem they address. An agent's skill library is a set of \texttt{SKILL.md} files, each a capability described by a name and a one-line description in its frontmatter, plus metadata (department, category, tags, and hand-authored related-skill references). The library is reached through a gateway of meta-tools rather than by loading files directly, and there is no ``load all skills'' path: an agent would otherwise have to pull every skill's discovery row into context, and that is the baseline both our systems improve on.

Two access routes define the setting. In the \emph{pull} route, the agent issues a query and receives a ranked list of candidate skills, resolving only those it needs; this is the route the ranker (\S\ref{sec:ranker}) serves and the one this paper evaluates. A second \emph{push} route, in which a per-turn hook injects an advisory list of candidate skills without loading them, exists in the same system but is not evaluated here. Both routes are query-conditioned: they answer ``which skill matches this request?'' Neither answers ``given the skill I am now using, what comes next?'', the gap the knowledge graph (\S\ref{sec:approach}) is designed to fill, and the distinction the rest of the paper turns on.

The paper follows one line of argument. We present the knowledge graph and its edge-generation pipeline (\S\ref{sec:approach}), then the hybrid ranker it is measured against (\S\ref{sec:ranker}), then the head-to-head comparison and its explanation (\S\ref{sec:results}), and finally what that explanation implies (\S\ref{sec:discussion}). The thread connecting these sections is a single claim in two parts: retrieval is not solved, and the graph nonetheless does not improve it, for a structural reason we make precise.

\paragraph{Contributions.}
\begin{enumerate}\itemsep2pt
\item \textbf{The pre-filter topology bound} (\S\ref{sec:topology}): a mechanistic, reproducible result that drawing a knowledge graph's candidate edges from an embedding's nearest-neighbours caps the graph at that embedding's own topology; the language model can add semantics to an edge but cannot extend the graph's reach.
\item \textbf{Five converging measurements} (\S\ref{sec:arms}): at matched token budget, substituting graph neighbours for additional ranked results is significantly worse ($p = 0.0007$); the LLM edge layer adds nothing over free embedding neighbours; $73\%$ of missed queries are unreachable through the graph; search recovers $2{\times}$ more often where the entry guess is wrong.
\item \textbf{Author-written query sets inflate retrieval accuracy} (\S\ref{sec:inflation}): the same ranker and catalogue score $44$ points higher (keyword) and $21$ points higher (hybrid) on an author-written query set than on a non-echo set, a transferable methodological result.
\item \textbf{The method and its cost} (\S\ref{sec:method}): LLM-at-add-time typed-edge generation, producing $1421$ typed edges for ${\sim}\$2.70$, hash-cached, cycle-safe, provenance-tagged, and reproducible, with an exhaustive $8.5\%$ direction-error census. The method works; the ranker is simply the better fit for entry retrieval.
\end{enumerate}

\section{Approach: A Typed Knowledge Graph}
\label{sec:approach}

 We describe the graph's edge model and its edge-generating pipeline and report the generator's coverage, cost, and edge precision.

\subsection{The edge model}
\label{sec:model}

The graph has three edge layers, which differ in provenance and in how far we trust them. \textbf{Curated} edges are meaningful workflow relationships, hand-authored or LLM-generated. \textbf{Derived} edges come from each skill's \texttt{related} metadata. \textbf{Inferred} \texttt{similar\_to} edges are embedding nearest-neighbours, the automatic backbone.

The curated layer has eight typed relations. Each carries a traversal priority rather than a continuous weight: \texttt{requires} ($\to$, prereq), \texttt{specializes} ($\to$, IS-A), \texttt{feeds\_into} ($\to$, data flow), \texttt{routes\_to} ($\to$, hand-off), \texttt{precedes} ($\to$, soft order), \texttt{complements} ($\leftrightarrow$), \texttt{alternative\_to} ($\leftrightarrow$), and \texttt{verifies} ($\to$). An author writes one direction. The runtime indexes both endpoints, so the edge set never doubles. That rule avoids hand-mirroring every edge, an $O(N^2)$ burden that stalled an earlier Go-based attempt after wiring only $16$ of $134$ skills. Four types (\texttt{requires}, \texttt{specializes}, \texttt{feeds\_into}, \texttt{precedes}) must stay acyclic.

In the current corpus, the curated and derived layers are both empty: only $1$ of $689$ skills declares \texttt{related} metadata, and no hand-authored edges exist. All active edges are LLM-generated ($1421$) or embedding-inferred ($1626$). The design ranks curated above \texttt{llm}, so a hand-authored edge overrides a model's guess; this is the mechanism that licences shipping the LLM layer at all. With zero curated edges, that override is implemented but unexercised.

The \texttt{similar\_to} backbone links each skill to its nearest neighbours in embedding space. Write $\hat{e}_a$ for the L2-normalised embedding of skill $a$. Cosine similarity is then the dot product
\begin{equation}
\operatorname{sim}(a,b) = \hat{e}_a \cdot \hat{e}_b .
\end{equation}
The neighbourhood $N_K(a)$ of skill $a$ is the set of its $K{=}8$ highest-similarity neighbours whose similarity also meets a threshold $\tau = 0.5$:
\begin{equation}
N_K(a) = \operatorname*{top\text{-}}K_{\,b \neq a}\ \{\, b : \operatorname{sim}(a,b) \ge \tau \,\},
\end{equation}
yielding $1626$ pairs over $690$ nodes. Only id-pairs enter the committed graph. The scores stay in a seed file, so floating-point noise never churns the graph.

\subsection{Edge generation}
\label{sec:method}

\paragraph{Why not learn edges from usage first?} The best evidence that two skills are related is that people use them together. That was our first approach: read the organisation's version-control history, find sequences where one skill follows another, and draw an edge wherever a pairing recurs. Two problems made it impractical. Counting co-occurrence needs one place where all the sequences are logged, and none existed yet. And usage only shows what follows what, which suffices for \texttt{precedes} and \texttt{complements}, but a sequence never reveals that $A$ \emph{requires} $B$, or that $A$ is a \emph{special case of} $B$. A new skill also has no history to learn from. We therefore switched to asking a model when a skill is added, which needs no history and works from the first skill. We return to the usage approach in \S\ref{sec:future}: the missing sink has since shipped, and it is retained in the design as a future \texttt{learned} provenance.

\paragraph{Coarse-to-fine generation.} When a skill is added, we first take its top-$K{=}8$ embedding neighbours; that is the coarse step. We then make one call to an inexpensive model (\texttt{claude-haiku-4-5}) under forced tool use. It returns typed edges through an \texttt{emit\_edges} tool whose schema pins each type to the curated vocabulary; that is the fine step. We keep only curated types between target and candidate, drop self-loops, and deduplicate. The neighbours are drawn across all departments, preserving cross-domain pivots while bounding cost to a constant $K$.

\paragraph{Incremental cost.} Each skill is keyed by \texttt{sha256(description)[:16]}. Let $\Delta$ denote a set of skills added or edited, $s$ a skill, $\operatorname{hash}(s)$ its current description hash, and $\operatorname{hash}_{\text{prev}}(s)$ the hash recorded in the previous build artifact. The generator runs only for skills whose hash has changed, so the number of model calls is the count
\begin{equation}
\operatorname{calls}(\Delta) = \bigl|\{\, s : \operatorname{hash}(s) \neq \operatorname{hash}_{\text{prev}}(s)\,\}\bigr|,
\end{equation}
and adding one skill costs one call rather than a full rebuild. An edge depends on both endpoints, but we key it on only one. It can therefore hold a stale view of its far endpoint until that skill is itself reprocessed. The edges are advisory, and they heal on the next build.

\paragraph{Cycle-safety, provenance, determinism.} A model can hallucinate a cycle, so we run a depth-first search and drop the back edges, separately for each acyclic type. Every edge carries provenance (empty for hand-authored, \texttt{llm}, or a reserved \texttt{learned}). Only one step is non-deterministic, the LLM call. It writes its output to an artifact, and the build then consumes that artifact deterministically, exactly as it consumes the embedding seeds.

\paragraph{Build results (corpus $=690$).} The generator produced $1421$ typed edges with zero failed API calls in ${\sim}15.5$ minutes; validation dropped $113$ hallucinated cyclic edges, which keeps the acyclic types acyclic. Type distribution: \texttt{complements} $390$, \texttt{feeds\_into} $350$, \texttt{precedes} $242$, \texttt{routes\_to} $128$, \texttt{requires} $108$, \texttt{alternative\_to} $101$, \texttt{specializes} $89$, \texttt{verifies} $13$. (The shortlists came from a $639$-era seed file, so about $50$ newer skills received no edges. A refresh is pending; it will change coverage, not the results here.)

\begin{figure*}[h]
\centering
\includegraphics[width=0.78\textwidth]{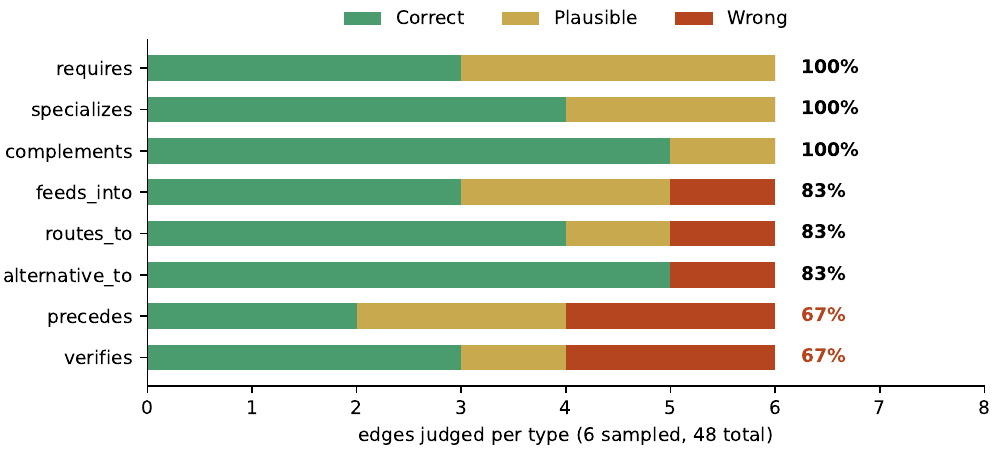}
\caption{\textbf{Most edge types are accurate; direction-based types miss more.} Per-type audit verdicts ($n{=}48$, six per type); percentages are the correct-or-plausible share.}
\label{fig:prec}
\end{figure*}
\paragraph{Cost.} At $\$1$/M input and $\$5$/M output tokens, the full backfill cost ${\sim}\$2.70$, below the ${\sim}\$3$ estimate; incremental cost is ${\sim}\$0.005$ per edited skill. As \S\ref{sec:arms} will show, this expenditure buys edges that add nothing over neighbours obtained free from a local embedding pass.

\paragraph{Precision.} Because edges are \emph{predicted}, not observed, we audited $48$ ($6$ per type), judging each from both endpoints as Correct/Plausible/Wrong. Loose precision (C+P) is $41/48$ (${\sim}85\%$); strict (C only) is $29/48$ ($60\%$). Those aggregates average over two populations that behave differently. Figure~\ref{fig:prec} separates them by relation type, six edges each. Read that way, the three relations whose meaning does not depend on which endpoint comes first (\texttt{requires}, \texttt{specializes}, \texttt{complements}) are entirely correct-or-plausible, while the three whose meaning does (\texttt{precedes}, \texttt{verifies}, \texttt{routes\_to}) fall to $67$--$83\%$, with \texttt{feeds\_into} and \texttt{alternative\_to} in between. The single-number precision therefore understates the safe types and overstates the directional ones, and the split is ordered cleanly enough by directionality to suggest a systematic error mode rather than sampling noise. \S\ref{sec:census} confirms this at population scale, where every contradiction category turns out to be a direction disagreement.

\section{Skill Retrieval by Ranking}
\label{sec:ranker}

Having built the graph, we turn to the other product of this work, the hybrid ranker that serves the pull route introduced in Section~\ref{sec:background}. It is the baseline we measure the graph against. This section describes the ranker and the efficiency of sparse loading; the retrieval accuracy it achieves, and the headroom it leaves, are reported in \S\ref{sec:results}.

\paragraph{What the ranker scores.} The ranker scores catalogue entries of \texttt{\{id, kind, name, description, path\}}. It never returns \texttt{path}, so it scores each capability on \texttt{name}~+~\texttt{description} alone: one line of frontmatter. As noted in Section~\ref{sec:dataset}, the $690$ full skill \emph{bodies} are indexed by neither the ranker nor the edge generator; a sibling effort found that indexing bodies degrades retrieval under this short-context encoder.

\paragraph{Three rankers.}\textbf{Keyword} (BM25)~\citep{robertson2009bm25} is dependency-free and always available, strong on exact identifiers and rare tokens, and serves as an off-topic gate. \textbf{Semantic} uses MiniLM vector similarity and is strong on paraphrase. \textbf{Hybrid} fuses the two and is our production default. The embedding model is \texttt{all-MiniLM-L6-v2}~\citep{reimers2019sentence} at a pinned revision: $384$-dim, mean-pooled, L2-normalised so cosine equals the dot product. It is \emph{the same model and revision the graph's \texttt{similar\_to} layer uses}, and the topology bound (\S\ref{sec:topology}) turns out to depend on that. This shared embedding is not incidental: both systems rest on the same similarity signal, which is exactly why one bounds the other.

\subsection{The efficiency win, and the question it raises}
\label{sec:win}

\begin{figure}[h]
\centering
\includegraphics[width=\linewidth]{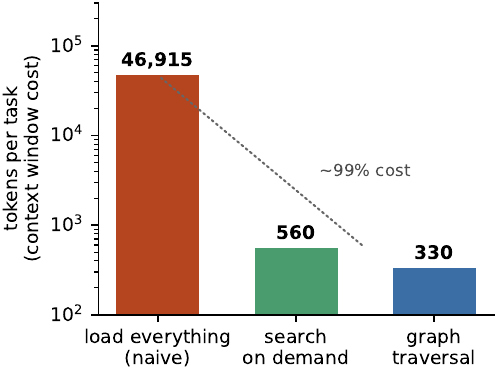}
\caption{Tokens per task ($690$ nodes), logarithmic axis.}
\label{fig:cost}
\end{figure}

The payoff of sparse ranked loading is large and direct. The naive load-all harness costs $46{,}915$ tokens per task ($875$ catalogue rows), while retrieving on demand costs ${\sim}560$, a $98.8\%$ reduction. Figure~\ref{fig:cost} compares all three regimes on one logarithmic axis. Search on demand and graph traversal cost almost the same, and both cost two orders of magnitude less than loading everything, so the saving comes from loading less rather than from loading smarter. Paired with hit@5 of $0.735$ on realistic queries (\S\ref{sec:negative}), the ranker is a deployable improvement over the naive baseline on its own, though it leaves real room for improvement.

That efficiency sharpens the second question. Sparse loading solves cost, but the ranker still misses roughly a quarter of realistic queries, so there is retrieval headroom for added structure to capture. The knowledge graph encodes workflow relations a ranker does not represent, and the rest of the paper tests whether that structure closes the gap.

\section{Dataset}
\label{sec:dataset}

All measurements are taken on one skill library and one evaluation.

\paragraph{Corpus.} The library contains $690$ skills across five departments at the time of our edge-generation run; earlier retrieval measurements were taken at $639$ skills, as the corpus grows by daily import. We state the corpus size at each measurement rather than as a single figure. Each skill contributes one node; department, category, and tags are node attributes, not edges. The full skill \emph{bodies} are not indexed by either system; the ranker and the edge generator both read only the one-line \texttt{description}. A sibling effort tested body indexing separately and found it degrades retrieval under this short-context encoder, though contemporaneous work at larger scale reports the opposite~\citep{skillrouter}; we treat this direction as open (\S\ref{sec:future}).

\paragraph{Evaluation sets.} We measure retrieval on two query sets, and the contrast between them is itself a finding (\S\ref{sec:inflation}). The primary set holds $117$ natural-language queries. Each maps to a verified expected skill and is tagged \emph{direct}, \emph{paraphrase}, or \emph{indirect}, and an integrity test rejects any query that echoes a skill's description. We use it for every headline retrieval number, and we credit the sibling effort that built it. The second is a $37$-query set written by the authors while reading the corpus, reported only to quantify the inflation that author-written queries introduce. Edge precision is measured on a separate sample of $48$ generated edges, six per type (\S\ref{sec:results}).

\section{Experiments and Results}
\label{sec:results}

We now compare the two systems. We first report retrieval accuracy (\S\ref{sec:negative}) and the inflation an author-written query set would have introduced (\S\ref{sec:inflation}). We then present the pre-filter topology bound that structurally predicts the graph's failure (\S\ref{sec:topology}), and finally test five graph usage modes against that prediction (\S\ref{sec:arms}).

\subsection{Retrieval accuracy}
\label{sec:negative}

\begin{table}[h]
\centering
\caption{Retrieval on the $117$-query non-echo set. Hybrid leaves $26.5\%$ of queries unserved at top-5.}
\label{tab:retrieval}
\setlength{\tabcolsep}{4pt}
\begin{tabular}{lccc}
\toprule
Ranker & top-1 & top-5 & indirect top-5 \\
\midrule
keyword (BM25) & 0.274 & 0.504 & 0.419 \\
\textbf{hybrid (default)} & \textbf{0.504} & \textbf{0.735} & \textbf{0.628} \\
\bottomrule
\end{tabular}
\end{table}

We evaluate on the retrieval set:\footnote{The $117$-query non-echo evaluation set was constructed by a sibling team within the organisation. It is included in the repository; we did not control its construction but verified its integrity test.} 117 task-to-skill cases across roughly 15 domains, a catalogue of 875 entries, $k{=}5$. Its authors worded the queries deliberately not to echo the target skill's description, tagged each \emph{direct}, \emph{paraphrase}, or \emph{indirect}, and added an integrity test that rejects echoing queries. Table~\ref{tab:retrieval} reports the result. The production hybrid ranker retrieves the correct skill into the top five for $73.5\%$ of queries, and for only $62.8\%$ when the user describes a symptom rather than naming the skill. Retrieval is therefore not solved: roughly $31$ of $117$ queries are still yet to be answered.

\subsection{Author-written query sets inflate retrieval accuracy}
\label{sec:inflation}

\begin{figure}[h]
\centering
\includegraphics[width=\linewidth]{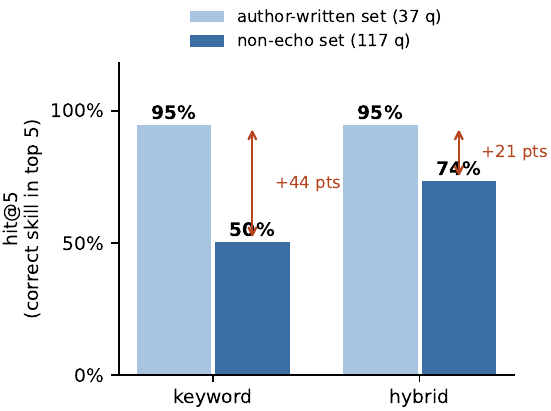}
\caption{\textbf{Author-written queries inflate retrieval accuracy.} hit@5 on the author-written ($37$-query) and non-echo ($117$-query) sets; only the queries differ.}
\label{fig:retrieval}
\end{figure}

The graph-boost result depends on measuring retrieval accurately, which proved to be the harder problem. An earlier version of this work used a $37$-query set written by the same author who built the system. Evaluated with the same ranker and catalogue, that set reports hit@5 of $0.946$ for both keyword and hybrid, far above the non-echo figures of $0.504$ and $0.735$. The gap is $44.2$ points for keyword and $21.1$ points for hybrid. Figure~\ref{fig:retrieval} sets the two query sets side by side for each ranker. The gaps are unequal enough to change the ordering: on the author-written set the two rankers are level, and on the non-echo set they separate by more than $23$ points. Only the queries change between the two conditions, so an evaluation built the natural way would have reported not merely inflated accuracy but a false equivalence between a lexical baseline and the production ranker.

The mechanism is description echo. Queries written while reading the corpus reuse the vocabulary of the descriptions being searched, and because BM25 scores exact term overlap, an echoing query approximates a primary-key lookup. This is why keyword collapses hardest when the echo is removed. We report both query sets rather than silently adopting the non-echo one, because the delta is itself the finding: a naive author-written evaluation would have reported near-perfect retrieval and hidden the headroom the graph then fails to capture.

\subsection{The pre-filter topology bound}
\label{sec:topology}

This is the paper's core structural finding, and we present it \emph{before} the graph usage experiments because it predicts their outcome. The coarse-to-fine design was presented as a \emph{cost} bound; it is also a \emph{topology} bound, not previously recognised.

\begin{figure}[h]
\centering
\includegraphics[width=\linewidth]{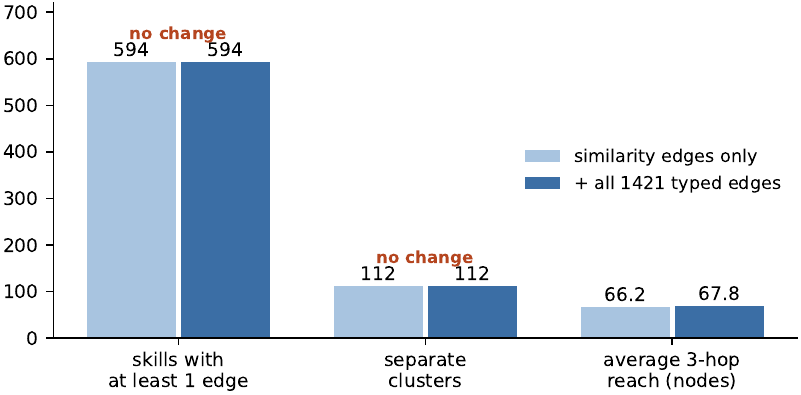}
\caption{Connectivity of the inferred backbone before and after adding all $1421$ typed edges.}
\label{fig:topo}
\end{figure}

\paragraph{Typed edges do not change reachability.} Adding all $1421$ typed edges to the inferred backbone leaves the graph's connectivity essentially unchanged. Table~\ref{tab:layers} gives the three layer sets side by side; the last two rows are the comparison that matters. Adding all $1421$ typed edges to \texttt{similar\_to} leaves wired nodes at $594$ (\textbf{zero} newly connected) and components at $112$ (\textbf{zero} merged). The largest component is unchanged, and 3-hop mean reach rises $2.4\%$. Figure~\ref{fig:topo} shows the same three metrics before and after. An entire layer of $1421$ edges is added, and the structure it was supposed to create does not appear.

\begin{table}[h]
\centering
\caption{Graph structure by layer set (reach over wired nodes, undirected). Adding the typed layer to \texttt{similar\_to} changes no connectivity metric.}
\label{tab:layers}
\setlength{\tabcolsep}{3pt}
\begin{tabular}{lccccc}
\toprule
layer set & wired & comps & largest & singles & 3-hop \\
\midrule
typed only & 524/690 & 190 & 441 & 166 & 29.6 \\
\texttt{similar\_to} & 594/690 & 112 & 544 & 96 & 66.2 \\
\textbf{all layers} & \textbf{594/690} & \textbf{112} & \textbf{544} & \textbf{96} & \textbf{67.8} \\
\bottomrule
\end{tabular}
\end{table}

\begin{figure}[h]
\centering
\includegraphics[width=\linewidth]{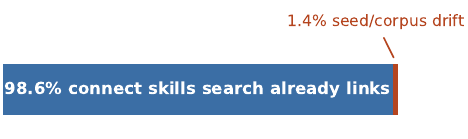}
\caption{Share of the $1022$ distinct typed pairs that also appear as \texttt{similar\_to} pairs.}
\label{fig:overlap}
\end{figure}

\paragraph{The mechanism, measured.} Of $1022$ distinct typed pairs, $1008$ ($98.6\%$) are \emph{also} \texttt{similar\_to} pairs; only $14$ ($1.4\%$) fall outside, attributable to seed/corpus drift (inference, not measurement). Figure~\ref{fig:overlap} shows that split. The $1.4\%$ is not novel structure the model found; it is drift between the seed file and the current corpus. Because the candidate shortlist \emph{is} the embedding top-$K$, a typed edge can only ever be proposed between skills the embedding layer already connected.

\begin{keybox}
\textbf{Topology bound (general property of the method class).} When LLM edge generation draws candidates from an embedding's top-$K$ neighbours, the resulting graph is confined to that embedding's topology. The LLM can enrich \emph{semantics} by assigning a typed relation to a pair, but it cannot add \emph{reach}, because it is never shown a pair the embedding did not already surface. Any method that pre-filters candidates by the same signal the ranker uses inherits this ceiling.
\end{keybox}

The topology bound predicts two things: re-ranking can only reshuffle within the ranker's own candidate set, so no retrieval improvement is mechanistically available; and graph neighbours at matched budget should underperform ranked results, because the graph draws from a subset of the same signal. The next section tests both predictions.

\subsection{Graph usage experiments}
\label{sec:arms}

\begin{figure}[h]
\centering
\includegraphics[width=\linewidth]{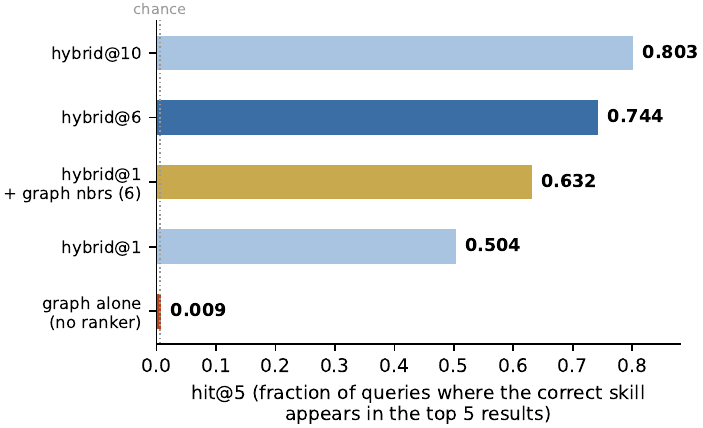}
\caption{Performance ladder. At matched budget (6 candidates), graph neighbours significantly underperform additional ranked results.}
\label{fig:arms}
\end{figure}

We test the graph in every plausible usage mode. Table~\ref{tab:arms} and Figure~\ref{fig:arms} report the full ladder; the graph cannot retrieve without a ranker, so every graph arm calls hybrid to pick its entry point.

\begin{table}[h]
\centering
\caption{Performance ladder on the $117$-query non-echo set. Every graph arm uses hybrid internally for entry selection.}
\label{tab:arms}
\setlength{\tabcolsep}{3pt}
\small
\begin{tabular}{lrl}
\toprule
arm & hit@5 & note \\
\midrule
graph alone (no ranker) & 0.009 & ${\approx}$chance \\
hybrid@1 & 0.504 & search, 1 cand \\
hybrid@1 + graph (6) & 0.632 & search+graph \\
\textbf{hybrid@6} & \textbf{0.744} & \textbf{search, 6 cands} \\
hybrid@10 & 0.803 & search, 10 cands \\
\bottomrule
\end{tabular}
\end{table}

\paragraph{Substitution at matched budget.} The apples-to-apples question is: \emph{given room for $N$ candidates, should they be $N$ ranked results or $1$ result plus $N{-}1$ neighbours?} At budget $6$, \texttt{hybrid@1+graph(6)} scores $0.632$ against \texttt{hybrid@6} at $0.744$, a deficit of $11.2$ points (McNemar $p = 0.0007$). The graph is significantly worse at matched budget on this corpus. The improvement over \texttt{hybrid@1} ($0.504 \to 0.632$) reflects the value of having \emph{any} additional candidates, not the value of graph structure.

\paragraph{The LLM edge layer adds nothing over free neighbours.} At budget $6$, varying only which edge layer supplies the neighbours: \texttt{similar\_to} alone (free local kNN) scores $0.632$; typed edges alone (the $1421$ LLM-generated edges) score $0.607$; both layers together score $0.632$. The entire LLM edge-generation step ($1421$ edges, ${\sim}\$2.70$, eight curated types) contributes zero measurable retrieval value over neighbours obtained free from a local embedding pass. This is the topology bound expressed as an outcome rather than a structural statistic.

\paragraph{73\% of missed queries are unreachable.} Of the $30$ queries \texttt{hybrid@6} misses, we measure reachability from hybrid's top-1 result through the graph: $6$ ($20\%$) are reachable within $1$ hop, $2$ ($7\%$) at $3$ hops, and $22$ ($73\%$) are unreachable within $3$ hops. The headroom is structurally out of the graph's reach at any ranking or edge quality; no amount of scoring improvement connects a target that is not connected.

\paragraph{Where the entry guess is wrong, search recovers twice as often.} Conditioning on whether hybrid's top-1 result is correct: when the entry is correct ($n{=}59$), both arms score $1.000$ trivially. When the entry is wrong ($n{=}58$), retrieving more ranked results recovers $48.3\%$ of cases while graph neighbours recover $25.9\%$. The graph underperforms precisely where help is needed.

\paragraph{Re-ranking (graph-boost) is a null.} For completeness, we also test re-ranking the ranker's output by spreading activation over the graph. Hybrid gains a single query in $117$ ($+0.85$ points; $6$ gained, $5$ lost; McNemar $p \approx 1.000$; $95\%$ interval $-4.8$ to $+6.5$ points). This is a null: no evidence of an effect of either sign.

\section{Discussion and Analysis}
\label{sec:discussion}

\subsection{Why ranking is better at every usage mode}
\label{sec:why}

Three mechanisms explain why five independent measurements converge.

First, \textbf{search is query-conditioned; the graph is not.} Search ranks $690$ capabilities against the user's actual task text; graph adjacency is fixed at build time. For ``select the right skill for \emph{this} task,'' search has strictly more information.

Second, \textbf{the topology bound (\S\ref{sec:topology}) confines the graph to a strict subset of the ranker's signal}. The LLM edge layer adds zero reach (Table~\ref{tab:layers}), zero retrieval value over free kNN (\S\ref{sec:arms}), and $73\%$ of the queries the ranker misses are not reachable through the graph at all. A truncated view of a signal cannot beat the signal.

Third, \textbf{hybrid fusion already covers both failure modes} the graph might have patched: the semantic channel covers paraphrase gaps and BM25 covers lexical drift, which is why keyword sheds $44$ points when echo is removed but hybrid sheds only $21$ (Figure~\ref{fig:retrieval}). The residual headroom that a ``graph rescues buried queries'' narrative would require has already been absorbed by the fusion.

\subsection{Statistical power: the null is a bound, not a shortfall}
\label{sec:power}

The substitution arm is significant ($p = 0.0007$); it is the re-ranking arm that requires care. Table~\ref{tab:retrieval} reports hit@5 of $0.735$; its $95\%$ half-width is $\pm 8.0$ points at $n{=}117$. On the re-ranking arm, the graph-boost difference is $+0.9$ points (Newcombe $95\%$ interval $-4.8$ to $+6.5$; exact McNemar $p = 1.000$). The interval excludes gains larger than roughly $6.5$ points, so re-ranking cannot recover more than about a quarter of the $26.5$-point headroom. What the null supplies is an upper bound on the effect, not a demonstration that the effect is exactly zero. Per-phrasing strata ($n{=}30$--$44$) are too small to support any sub-group conclusion; the mechanism, not the sample size, is what makes the aggregate null strong.

\subsection{Direction-error census}
\label{sec:census}

The direction-reversal mode visible per type in Figure~\ref{fig:prec} is confirmed at population scale. Of $1022$ distinct pairs, $298$ ($29.2\%$) carry more than one type. Of those, $87$ ($8.5\%$ of all pairs) are \emph{contradictory}. Twenty carry the same asymmetric type in both directions, which is an unambiguous error needing no semantic judgement. Every contradiction category is a direction disagreement (e.g.\ $20$ \texttt{routes\_to}-vs-\texttt{routes\_to} reversed; $17$ \texttt{feeds\_into}-vs-\texttt{precedes} reversed), so the $48$-edge audit and the $1022$-pair census agree on which relations fail and on how they fail. The implication for any product use is concrete: shipping ``do $X$ before $Y$'' guidance at $8.5\%$ reversed is worse than shipping nothing, because a wrong prerequisite actively misleads.

\subsection{When the knowledge graph would be more beneficial}
\label{sec:when}

Everything above concerns \emph{retrieval}. The graph's remaining structural case is on \emph{ordering}: the question of what skill follows the current one, which retrieval does not pose. Consider a task needing $n$ skills in sequence. Repeated search succeeds with probability $p^n$; graph traversal succeeds with $p\cdot q^{\,n-1}$, where $q$ is per-hop edge accuracy. The graph wins iff $q > p$. The directional edge types that an ordering consumer would read run $0.67$--$0.83$ accuracy, which straddles the $p{=}0.735$ threshold. Our numbers cannot settle whether traversal beats repeated search for ordering; the question requires a skill-to-skill sequence dataset that does not yet exist and an $8.5\%$ direction-error rate that is unmet. This is future work, not a rescue for entry retrieval.

The graph need not outperform search to be useful: enriching \texttt{resolve\_skill} so that an agent loading a skill also receives its prerequisites, next-steps, and alternatives is an additive change with no adoption cost, provided the direction-error rate is first reduced.

\subsection{Threats to validity}
\label{sec:threats}

We surface the threats a careful reviewer will raise. The primary evaluation set has $117$ queries; one query is ${\sim}0.85$pp of hit@5. The non-echo set and its integrity test were built by a sibling effort, not by us, which removes the author-echo bias but means we did not control its construction. The precision audit behind Figure~\ref{fig:prec} is $n{=}48$, single-judge, and description-only, which is why the $87$ structural contradictions (immune to this critique) carry the weight. No latency was measured. We never tested sufficiency: every cost arm counts rows rather than completed tasks. Our metric is single-gold hit@5; under it, a graph that supplies a needed prerequisite but displaces the gold skill from the top five scores as a loss. Finally, this is a single corpus, single organisation, single embedding model: no claim about other libraries, scales, or backends is supported.

\subsection{Future work}
\label{sec:future}

In priority order. \textbf{Co-usage telemetry:} an ordered skill-call log has shipped to production; observed co-usage is orthogonal to the embedding and is the one remaining candidate signal that is not embedding-confined. It could in principle break the topology bound. \textbf{The sequence and sufficiency experiment} is the load-bearing missing test. A harness exists that measures next-skill invocation with and without typed data; the decisive experiment is buildable but needs runtime wiring and must first resolve a known backfire effect from the suggestion hook. \textbf{Candidate generation beyond top-$K$} is the only route that could give the graph reach the ranker lacks. \textbf{A graph-native retrieval system:} the topology bound shows that an embedding-confined graph cannot extend retrieval reach; a system that generates candidate edges from sources beyond the embedding kNN (e.g.\ co-usage logs, dependency metadata, or cross-modal signals) could in principle break this ceiling and provide a retrieval mechanism that complements rather than duplicates the ranker. \textbf{Disambiguation at scale:} at larger corpus sizes where embedding neighbourhoods grow denser, clusters of near-identical skills may become harder for the ranker to distinguish; the disambiguation value of typed edges within such clusters warrants separate investigation. \textbf{Body indexing under a long-context encoder:} a sibling effort found that indexing bodies reduces retrieval under our $256$-word-piece encoder, but \citet{skillrouter} report the opposite at scale, so we treat this direction as reopened rather than closed.

\section{Related Work}
\label{sec:related}

Two contemporaneous systems address our exact setting, retrieval over a library of \texttt{SKILL.md} files, and reach positive conclusions where we reach a negative. \textbf{SkillRouter}~\citep{skillrouter} studies retrieval over ${\sim}80$K skills and reports that the full implementation body, not the frontmatter description, is the decisive signal: removing it degrades retrieval by $29$--$44$ points across every method they test. Their two-stage retrieve-and-rerank pipeline reaches $74.0\%$ top-1. \textbf{Graph-of-Skills} (GoS)~\citep{graphofskills} builds a typed graph over a \texttt{SKILL.md} library and retrieves a dependency-aware bundle via hybrid semantic--lexical seeding and reverse-weighted Personalized PageRank, reporting gains on SkillsBench and ALFWorld together with large token reductions.

The divergence from our result lies in candidate generation. GoS derives its edges from dependency structure in the skill packages, not from the retriever's own embedding neighbourhood. Our generator does, and \S\ref{sec:topology} shows the consequence: $98.6\%$ of typed pairs fall inside the \texttt{similar\_to} set the ranker already computes. More broadly, graph-augmented retrieval has shown gains in compositional and multi-hop settings~\citep{edge2024graphrag,sing}; our result is notable precisely because it holds in a non-saturated \emph{single-hop} regime, which the topology bound explains.

\section{Conclusion}
\label{sec:conclusion}

We set out to improve how an agent selects skills from a large library, and built two systems toward that end. Sparse ranked loading cuts context cost by roughly two orders of magnitude, a real and deployable gain. The introduction promised a two-part thread, and we close it here. \emph{Retrieval is not solved}: on a non-echo set of $117$ realistic queries the hybrid ranker reaches the correct skill for $73.5\%$ of cases, leaving roughly a quarter unfulfilled. \emph{The graph does not close the gap, for a structural reason.} Five measurements converge: at matched budget, graph neighbours are significantly worse than additional ranked results ($p = 0.0007$); the LLM edge layer adds nothing over free embedding neighbours; $73\%$ of missed queries are unreachable through the graph; search recovers twice as often where the entry guess is wrong; and re-ranking with the graph is a null. The structural reason is that $98.6\%$ of the graph's edges connect skills the embedding layer had already connected, so the pre-filter that makes edge generation affordable also bounds the graph's topology.

The method works: it is cheap, reproducible, and reasonably accurate. What does not hold is the premise that an embedding-confined graph improves on a strong ranker at entry retrieval. Establishing that structurally, and identifying candidate generation beyond the embedding kNN and co-usage telemetry as the directions that remain open, is the contribution.

\section*{Reproducibility}

Repository \texttt{praetorian-inc/palatine}, tag \texttt{skill-graph-paper-v2}
(commit \texttt{17223a95}). Every figure reported here reproduces from that tag
and from no other point in the repository: the work was developed as a stack of
six pull requests, each contributing only its own edge layer, and three defects
surfaced only when they were merged. Setup, environment pins, and each command
with its expected output are in \texttt{docs/skill-graph-reproduction.md} at that
tag. No measurement requires an API call.

Two caveats recorded there are worth surfacing. The repository's graph and gateway
CI workflows trigger only on pull requests targeting \texttt{main}; every branch in
the stack targeted a sibling, so \emph{neither suite ever ran on this work} and the
local runs are the sole evidence. And the committed gateway bundle must be rebuilt
(\texttt{make gateway-build}) before any retrieval measurement, since a stale
bundle causes the semantic and hybrid arms to skip silently rather than fail.

The $117$-case non-echo retrieval dataset, its phrasing taxonomy, and the
description-only vs.\ description-plus-body ablation are prior work by a sibling
effort and ship on \texttt{main}; only the graph arms are ours.

\bibliographystyle{plainnat}
\bibliography{refs}

\end{document}